\documentclass[letterpaper, 10 pt, conference]{ieeeconf}  

\usepackage{cite}
\usepackage{amsmath,amssymb,amsfonts}
\usepackage{algorithmic}
\usepackage{graphicx}
\usepackage{textcomp}
\usepackage{xcolor}
\usepackage{yhmath}
\def\BibTeX{{\rm B\kern-.05em{\sc i\kern-.025em b}\kern-.08em
    T\kern-.1667em\lower.7ex\hbox{E}\kern-.125emX}}

\usepackage{array}
\usepackage[ruled,linesnumbered]{algorithm2e}

\usepackage{textcomp}
\usepackage{stfloats}
\usepackage{url}
\usepackage{verbatim}
\usepackage{graphicx}
\usepackage{cite}
\usepackage{subfigure}
\usepackage{caption}
\usepackage{multirow}
\usepackage[colorlinks,
            linkcolor=blue,
            anchorcolor=blue,
            citecolor=blue,
            urlcolor=blue]{hyperref}

\IEEEoverridecommandlockouts                              

\title{\LARGE \bf
NavG: Risk-Aware Navigation in Crowded Environments Based on Reinforcement Learning with Guidance Points
}

\author{Qianyi Zhang$^1$, Wentao Luo$^2$, Boyi Liu$^3$, Ziyang Zhang$^2$, Yaoyuan Wang$^2$, and Jingtai Liu$^{1,\dagger}$
\thanks{$^1$ Institute of Robotics and Automatic Information System, Nankai University, Tianjin Key Laboratory of Intelligent Robotics, Tianjin, China.} 
\thanks{$^2$ Advanced Computing and Storage Lab, Huawei 2012 Lab.} 
\thanks{$^3$ Department of Electronic and Computer Engineering, The Kong Kong University of Science and Technology. }
\thanks{$^{\dagger}$ Corresponding author with email: liujt@nankai.edu.cn. This work was done during Qianyi's internship at Huawei and is supported by National Natural Science Foundation of China under Grant 62173189.}
}

\begin{document}

\maketitle
\thispagestyle{empty}
\pagestyle{empty}

\begin{abstract}
Motion planning in navigation systems is highly susceptible to upstream perceptual errors, particularly in human detection and tracking. To mitigate this issue, the concept of guidance points—a novel directional cue within a reinforcement learning-based framework—is introduced. A structured method for identifying guidance points is developed, consisting of obstacle boundary extraction, potential guidance point detection, and redundancy elimination. To integrate guidance points into the navigation pipeline, a perception-to-planning mapping strategy is proposed, unifying guidance points with other perceptual inputs and enabling the RL agent to effectively leverage the complementary relationships among raw laser data, human detection and tracking, and guidance points.
Qualitative and quantitative simulations demonstrate that the proposed approach achieves the highest success rate and near-optimal travel times, greatly improving both safety and efficiency. Furthermore, real-world experiments in dynamic corridors and lobbies validate the robot’s ability to confidently navigate around obstacles and robustly avoid pedestrians.
\end{abstract}

\section{Introduction} \label{Int}
\begin{figure}
  \includegraphics[width=1.0\linewidth]{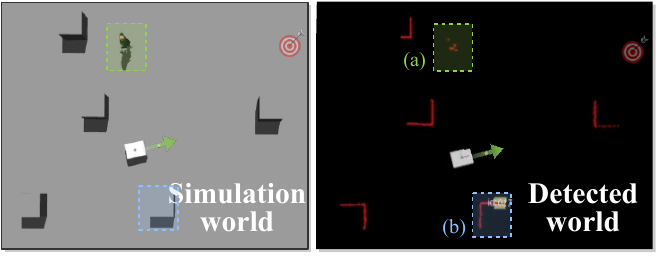}
	\caption{Issues in human detection. (a) The human is either not detected or detected with a velocity estimation error. (b) A non-human object is mistakenly classified as a human. Visit this \href{https://navg-dev.github.io}{website} for the performance of the proposed NavG.}
	\label{fig_problem}
\end{figure}

With the continuous advancement of robotic technologies, a widely accepted navigation framework has emerged, encompassing perception, planning, control, and localization~\cite{ref_review1, ref_review2}. As a downstream component, the planning module processes outputs from the perception module, such as segmented objects and detected pedestrians. Consequently, errors originating in perception inevitably propagate to planning. In particular, inaccuracies in human detection and tracking—including misestimating a pedestrian’s velocity, failing to detect a pedestrian, or misclassifying a non-pedestrian as a pedestrian, as illustrated in Fig.\ref{fig_problem}—can significantly compromise navigation safety and efficiency.

While researchers focused on human detection and tracking strive to improve accuracy, the inherent limitations of neural networks and the uncertainty of human behavior render perfect detection and tracking unattainable~\cite{ref_obj_detection1, ref_obj_detection2, sam}. As a result, risk-aware navigation, which necessitates the planning module to be able to perform robust planning itself in the presence of perception errors, has become a critical research topic~\cite{evora, drlvo}, which is also the focus of this paper.

In recent years, a verity of methods have been proposed to address risk-aware navigation. Some mathematical modeling-based approaches ensure safety by allocating an additional safety margin for the robot based on human-robot interaction~\cite{khambhaita2020viewing, stcteb}, human gaze information~\cite{10161222}, social norms~\cite{9775638}, or a speed map~\cite{9982200}, often at the cost of efficiency. Some supervised deep learning methods generate human-like trajectories by learning from labeled training data~\cite{evora, AgileButSafe, 10610665}. Meanwhile, deep reinforcement learning (DRL)-based methods enable the robot to learn autonomously through extensive interaction with its environment~\cite{hirose2024selfi, drlvo, ref_2_3}. Since neural networks inherently capture uncertainty and human-controlled movement is not always optimal, DRL has emerged as a promising approach, often demonstrating superior performance over alternative methods.

In reinforcement learning-based navigation, one prominent branch is to incorporate estimated human states, such as locations, velocities, and shapes, into the state function. CADRL~\cite{ref_1} was the first to apply deep reinforcement learning to navigation, using detected pedestrians as input and employing Q-learning to navigate in simple pedestrian-only scenarios. Building on this foundation, SA-CADRL~\cite{ref_2} and ssDRL~\cite{ref_3} introduce human social norms and social tension spaces, respectively, to enforce socially compliant and adaptive robot navigation.
To address the limitation of presetting a maximum number of pedestrians in network models, GA3C-CADRL~\cite{ref_4, ref_5} integrates an LSTM structure, theoretically allowing the RL agent to handle an arbitrary number of pedestrians. SARL~\cite{ref_2_3} further improves navigation by incorporating a self-attention mechanism to account for group interactions and dynamic movement patterns. Meanwhile, Thomas~\cite{ref_9} explores RL in scenarios where robots actively push movable obstacles to reach their destination efficiently. ViPlanner~\cite{roth2024viplannervisualsemanticimperative} enhances environmental understanding by explicitly incorporating object segmentation into the navigation process to improve flexibility.

\begin{figure*}[htb]
	\centering
    \includegraphics[width=1.0\linewidth]{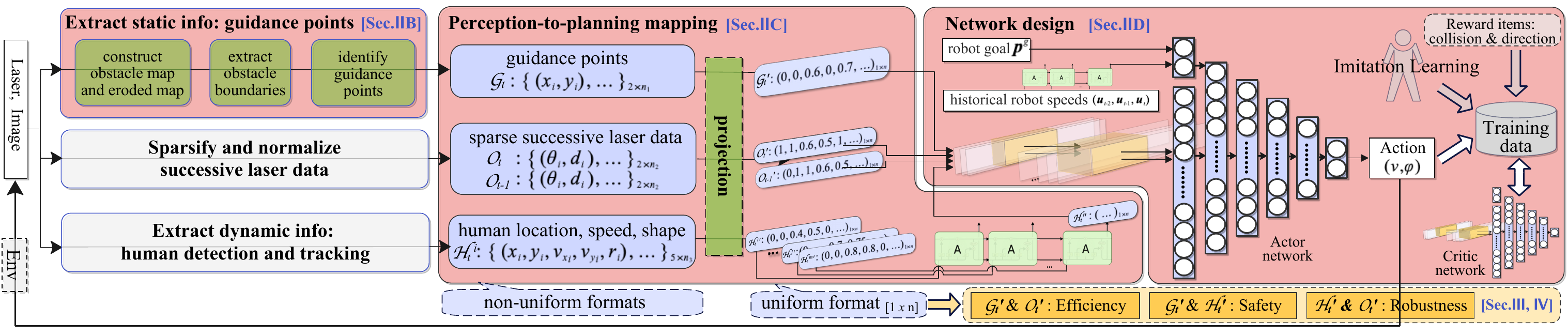}
	\caption{Framework of the proposed NavG. Raw sensor data is processed to extract sparse laser data, human states, and guidance points, which indicate potential directions for the robot. These elements are unified in polar coordinates centered on the robot. Along with the robot’s goal and historical states, all inputs are fed into the neural network to generate actions.
    }
    \label{fig_framework}
\end{figure*}

Another widely studied branch involves directly inputting raw sensor data, such as images or laser scans, into reinforcement learning models. Given the large state space, some studies integrate RL agents with model-based motion planners to enhance decision-making and efficiency. V.P.~\cite{ref_21} employs an RL agent to select a local target while relying on standard control methods for execution. LAMP~\cite{ref_19} introduces a hybrid navigation framework that dynamically switches between PID control and RL-based decision-making based on environmental characteristics. APPL~\cite{ref_20} incorporates user feedback to fine-tune the RL model’s parameters, improving adaptability.
GDAE~\cite{ref_8} integrates laser data and sub-targets to facilitate exploration in unknown environments. TERP~\cite{ref_16} utilizes 2D normalized elevation maps, robot poses, and target locations to find safe trajectories in uneven terrains. Swift~\cite{ref_22} applies RL to first-person drone racing, optimizing real-time maneuverability. DRL-VO~\cite{drlvo} combines RL with velocity obstacles, constraining the robot’s velocity to enable efficient and safe goal-directed navigation.



Building upon these works, this paper introduces the concept of guidance points in Sec.\ref{sec31}, which explicitly provide potential directions for the robot. At a macro level, guidance points help the robot recognize that it can safely pass behind a human rather than taking a riskier detour in front. At a micro level, they serve as an attractive force, preventing the robot from getting too close to humans or obstacles.
Guidance points complement raw sensor data and detected human states. Although these data sources vary in type and dimensionality, a perception-to-planning mapping strategy is introduced in Sec.\ref{sec32} to integrate them effectively, adaptively balancing navigation efficiency and safety. Furthermore, the complete navigation framework is presented in Sec.\ref{sec33}, demonstrating how these components work together to ensure safe and reliable navigation.

In summary, this work presents an RL-based navigation, as illustrated in Fig.\ref{fig_framework}. The main contributions are as follows:
\begin{itemize}
  \item A guidance point identification approach that robustly provides potential directions for the robot.
  \item A perception-to-planning mapping strategy to efficiently unify various perceptual information. 
  \item A navigation framework that is robust for perceptual errors in complex dynamic scenarios. 
\end{itemize}

\section{Methodology} \label{Met}

\subsection{Problem Formulation under RL Framework}
The collision avoidance problem can be formulated as a sequential decision-making process under the reinforcement learning framework. 
The state at the current moment $t$ consists of the observable state and the estimated state $\mathcal{S}_t = [\mathcal{S}_t^o, \mathcal{S}_t^e]$. 
The observable state $\mathcal{S}_t^o=[\textbf{p}_t, \textbf{p}^g, \textbf{u}_t, \mathcal{O}_t]$ denotes the values that the robot can directly access, including its current position $\textbf{p}_t = (x_t, y_t)$, goal position $\textbf{p}^g$, current action $\textbf{u}_t = (v_t, \phi_t)$, and raw sensor data $\mathcal{O}_t$.
The estimated state $\mathcal{S}_t^e = [\mathcal{G}_t, \mathcal{H}_t ]$ refers to the values extracted from the raw sensor data, including the guidance points $\mathcal{G}_t$ that guide the robot to pass through dense obstacles, and the detected objects $\mathcal{H}_t$ that specifically indicates human beings in this work. 
For our formulation, a robot with Ackermann wheels is considered, so the expected action in the next moment consists of the rear wheel's speed and front wheel's steer angle $\textbf{u}_{t+1} = (v_{t+1}, \phi_{t+1})$. 
The objective is to develop a policy $\pi : (\mathcal{S}_t, \textbf{u}_t) \Rightarrow \textbf{u}_{t+1}$ to minimize the time to goal while avoiding collisions with static obstacles and dynamic pedestrians.



\begin{figure*}[htb]
	\centering
    \includegraphics[width=1.0\textwidth]{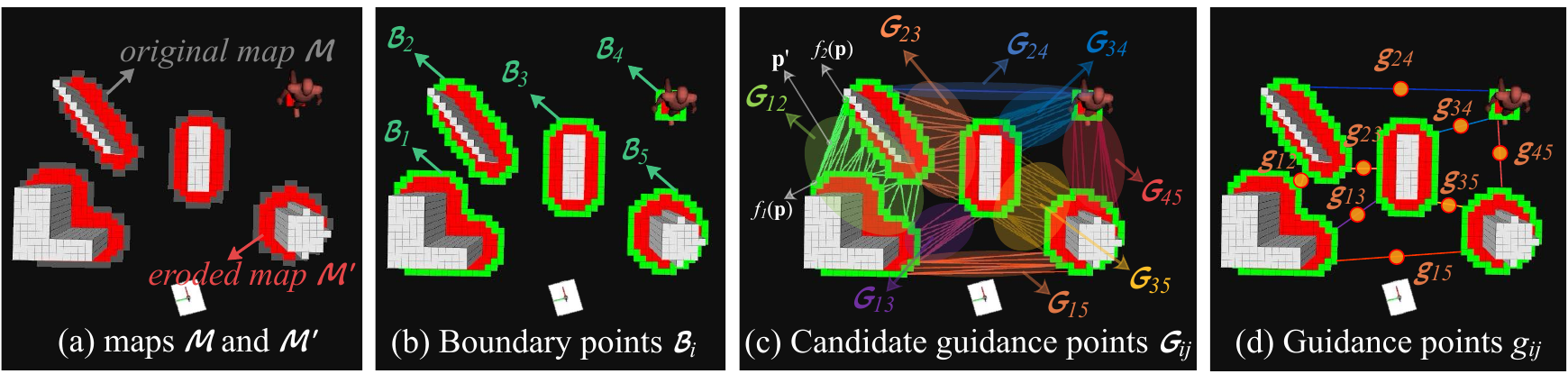}
	\caption{Illustration of guidance point identification. (a) Given an obstacle map, an erosion operation and depth-first search are applied to obtain (b) the eroded map and (c) boundary points. (d-e) Guidance points are the midpoints of boundary point pairs from different groups with the smallest distance.
    }
	\label{fig_guidancepoints}
\end{figure*}
\subsection{Identification of Guidance Points} \label{sec31}
The robot continuously observes the environment during its movement and maintains a 2D map $\mathcal{M}$, with $\mathcal{M}(\textbf{p})=1$ characterizing the presence of an obstacle at the point $\textbf{p}=(x,y)$ and $\mathcal{M}(\textbf{p})=0$ being a collision-free point. 
A single-step erosion operation is applied to $\mathcal{M}$, producing an eroded map $\mathcal{M}'$.
The intersection of the two maps, $\mathcal{M} \cap \mathcal{M}'$, allows the extraction of a set of boundary points $\mathcal{B}$: 
\begin{equation}
    \mathcal{B}=\{  \textbf{p}  |_{\mathcal{M}(\textbf{p})=1 \; \cap \; \mathcal{M}'(\textbf{p})=0}  \}
\end{equation}

By applying depth-first search to boundary points using a clockwise 8-connected neighborhood matrix, the algorithm clusters adjacent points into a boundary group $\mathcal{B}_i$, where the points are arranged in a clockwise order: 
\begin{equation}
	\begin{gathered}
            \mathcal{B} = \{ \mathcal{B}_1, \mathcal{B}_2, ..., \mathcal{B}_n \} \\ 
            \mathcal{B}_i = \{ p_j |_{\forall \textbf{p}_j \in \mathcal{B}_i \; \cap \; \forall \textbf{p}_{j+1} \in \mathcal{B}_i \; \cap \; ||\textbf{p}_j-\textbf{p}_{j+1}||_{\infty}=1 } \}
	\end{gathered}
\end{equation}

Taking boundary points as input, the method in \cite{ref_voronoi} is used to compute a distance map. 
For each collision-free point $\textbf{p}$ that has an equal minimum distance to two boundary points, $f_i(\textbf{p})$ and $f_j(\textbf{p})$, from different groups $\mathcal{B}_i$ and $\mathcal{B}_j$ respectively, the midpoint $\textbf{p}'$ of the line segment connecting $f_i(\textbf{p}) f_j(\textbf{p})$ is defined as the candidate guidance point.
\begin{equation}
    \mathcal{G}_{ij}=\{ \textbf{p}' |_{ \textbf{p}'=f_i(\textbf{p})/2+f_j(\textbf{p})/2 \; \cap \; f_i(\textbf{p}) \in \mathcal{B}_i \; \cap \; f_j(\textbf{p}) \in \mathcal{B}_j   } \}
\end{equation}

The final guidance point $\textbf{g}_{ij}$ between $\mathcal{B}_i$ and $\mathcal{B}_j$ is the point in $\mathcal{G}_{ij}$ that has the least distance to the boundaries. 
\begin{equation}
    \textbf{g}_{ij} = {\arg\min}_{\textbf{p}' \in \mathcal{G}_{ij}}  ||\textbf{p}'-f_i(\textbf{p})||_2 + ||\textbf{p}'-f_j(\textbf{p})||_2 \label{gij}
\end{equation}

\begin{figure}
  \includegraphics[width=1.0\linewidth]{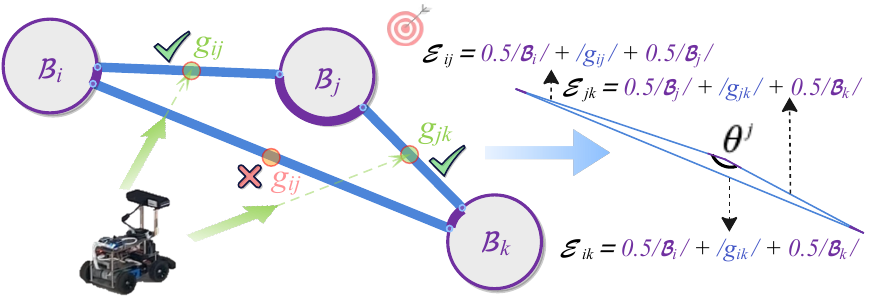}
	\caption{Illustration of eliminating the misleading guidance point $g_{ij}$ which may dangerously and redundantly guide the robot too close to $\mathcal{B}_j$. Removing it ensures efficient guidance to $g_{ij}$ or $g_{jk}$.}
	\label{fig_identification}
\end{figure}

To eliminate redundant guidance points that may even provide misleading directions, points that the robot cannot observe or that are positioned behind it are first removed.

Additionally, every three adjacent obstacle groups with IDs $i,j,k$ are concatenated to form a triangle (see Fig.\ref{fig_identification}). 
Defining $|g_{ij}|$ as the length of the line segment connecting two boundaries $\mathcal{B}_i$ and $\mathcal{B}_j$, and $|\mathcal{B}_i(g_{ij}, g_{ik})|$ as the arc length along $\mathcal{B}_i$, the length of a triangle edge $\mathcal{E}_{ij}$ is given by:
\begin{equation}
    \mathcal{E}_{ij}(g_{ij}) = |\mathcal{B}_i(g_{ij}, g_{ik})|/2 + |g_{ij}| + |\mathcal{B}_j(g_{ji}, g_{jk})|/2
\end{equation}

If the sum of any two edges exceeds the length of the remaining edge, or if the largest angle $\theta_i$ in the triangle exceeds a predefined threshold, the guidance point $g_{jk}$ located on the longest edge is removed. The remaining guidance points form the final list $\mathcal{G}_t$ at the current moment $t$.

\begin{figure*}[htb]
	\centering
    \includegraphics[width=1.0\linewidth]{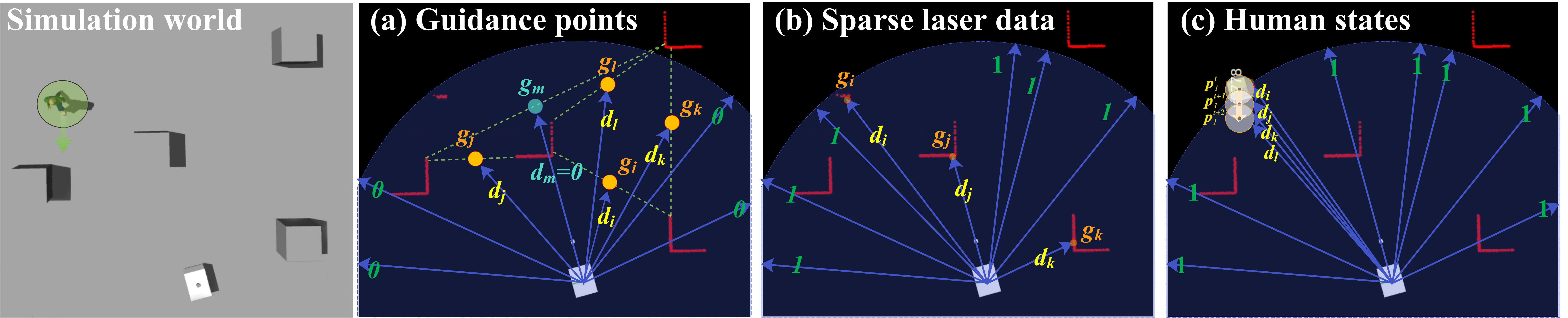}
	\caption{Illustration of projecting various data types of different lengths onto polar coordinates centered on the robot. The values are either positively or negatively correlated with their distances to the robot, ensuring consistency in how data magnitudes influence safety. Figures (a, b, c) share the relation $d_i > d_j > d_k > d_l$, indicating that (a) closer guidance points contribute more to safety, and (b-c) distant obstacles or pedestrians have a lower impact on collision risk for the current action. Their different combinations yield distinct benefits, which will be further analyzed in the simulation section.
    }
	\label{fig_perception2planning}
\end{figure*}

\subsection{Mapping from Perception to Planning} \label{sec32}
Recall that the perception module provides three types of information in different formats:
(a) Guidance points in the robot’s local Euclidean coordinate system, where each unit $\textbf{g}_{ij}=(x,y)$ (as defined in Eq.\ref{gij}) represents a location.
(b) Laser data in polar coordinates, where each unit $\textbf{o}_i=\{\theta, d\}$ represents the detected obstacle distance $d$ at angle $\theta$. 
(c) Detected human states in the global Euclidean coordinate system, where each unit $\textbf{h}_i=\{\textbf{p}, \textbf{v}, r\}$ consists of a human’s location $\textbf{p}=(x,y)$, velocity $\textbf{v}=(v_x,v_y)$, and a circular shape with radius $r$. 
In this section, these elements are unified into polar coordinates centered on the robot, forming the state space for the reinforcement learning framework.

\textbf{Guidance points} (Fig.\ref{fig_perception2planning}a). 
A guidance point list $\mathcal{G}_t'$ is initialized as a vector of $n$ zeros, indicating that no guidance points are available.  
Each guidance point $\textbf{g}_{ij} = (x,y)$ is projected into the robot's polar coordinate system and mapped to $\mathcal{G}_t'$. Its index $k$ is determined by its angle with respect to the robot, while its value $d_k$ is negatively correlated with its distance from the robot:
\begin{equation}
	\begin{gathered}
            k = \arctan(y,x) \times n/2\pi ,  \hspace{2em} d_k = 1 - ||\textbf{g}_{ij}||_2 / d_{\max}   \\
            \mathcal{G}_t' = [d_1, d_2, \ldots , d_n] 
	\end{gathered}
    \label{gtd}
\end{equation}
where $d_{\max}$ is the maximum detected distance.

\textbf{Sparse laser data} (Fig.\ref{fig_perception2planning}b). 
Given that neighboring points convey similar information and excessive data may hinder the convergence of the reinforcement learning network, we first sparsify the raw laser data $\mathcal{O}_t$ based on the resolution $\Delta \theta$, resulting in a reduced set $\mathcal{O}_t'$.
The remaining laser points are arranged in counterclockwise order within the polar coordinate system, ensuring that each adjacent point is separated by an angular difference of $\Delta \theta$.
Each point with index $i$ represents the maximum detected obstacle distance $d_i$ from the robot’s edge within the corresponding angular range.
Additionally, the sparse laser data from the previous time step, $\mathcal{O}_{t-1}'$, is incorporated to allow the Rl agent to infer the velocities of dynamic obstacles on its own.
\begin{equation}
	\begin{gathered}
            d_i = {\max}_{\theta \in [(i-0.5)\Delta \theta, (i+0.5)\Delta \theta]} \; \textbf{o}_i / d_{\max} \\
            \mathcal{O}_t' = [ d_1, d_2, \ldots , d_n ]
	\end{gathered}
\end{equation}
where $d_{\max}$ and $n$ are the same as those in Eq.\ref{gtd}.

\textbf{Human states} (Fig.\ref{fig_perception2planning}c). 
Similarly, the human state is projected onto the polar coordinate system. A list $\mathcal{H}_t'$ is initialized as a vector of $n$ zeros, indicating the absence of detected pedestrians.
For a pedestrian with ID $i$ at location $\textbf{p}_i$ with a radius of $r_i$, all points along its circular body shape are mapped into the polar coordinate system following the same principle used for guidance points and filled into $\mathcal{H}_t'$: 
\begin{equation}
    \mathcal{H}_t' = \{d_1, d_2, ..., d_n\}
\end{equation}

Furthermore, assuming the pedestrian moves according to the detected velocity, their future positions at time steps $t+1$ and $t+2$ are denoted as $\textbf{p}_i^{t+1}$ and $\textbf{p}_i^{t+2}$, respectively. Applying the same processing method as for $\textbf{p}_t$, the derived values are also projected onto $\mathcal{H}_t'$, where the minimum value is taken for the same index. 

In scenarios with a single pedestrian, this representation fully characterizes its presence. 
However, when multiple pedestrians are present, those processed later may overwrite information from those processed earlier, especially when their future positions are considered. This could introduce ambiguity in the RL state representation.
To address this, a separate list is first constructed for each pedestrian as described above.
Then, to distinguish between individuals, we temporarily denote the list for a pedestrian with ID $i$ as ${\mathcal{H}_t^i}'$.
To effectively model human interactions, LSTM~\cite{ref_lstm} is employed to aggregate the representations.
Given that pedestrians farther from the robot have a lesser impact on the robot's decisions—aligning with LSTM’s property of better memory retention for recent inputs—these human lists are fed into the LSTM in order from the farthest to the nearest, forming the final list $\mathcal{H}_t''$: 
\begin{equation}
    \mathcal{H}_t'' = LSTM({\mathcal{H}_t^i}', {\mathcal{H}_t^j}', ..., {\mathcal{H}_t^m}')
\end{equation}

\subsection{Framework of RL Navigation} \label{sec33}

Given the huge state space, Soft Actor-Critic (SAC)~\cite{ref_10, ref_16} is employed in this work. 
A critic network, $Q(\mathcal{S}_t, \textbf{u}_t)$, is trained to evaluate the agent's actions, while an actor network, $\pi (\mathcal{S}_t, \textbf{u}_t)$, generates the robot's actions with the assistance of the critic network. 
Updating both networks in parallel facilitates faster convergence. 

As illustrated in Fig.\ref{fig_framework}, both networks share the same input $[\mathcal{G}_t', \mathcal{O}_t', \mathcal{O}_{t-1}', \mathcal{H}_t'', \textbf{p}^{g}, (\textbf{u}_{t-2}, \textbf{u}_{t-1}, \textbf{u}_t)]$. The first four terms, previously introduced, share a uniform format of $[1 \times n]$ and are processed through two convolutional layers to extract relevant features.
$\textbf{p}^{g}$ represents the robot’s goal in the robot's polar coordinate, while the historical actions, $(\textbf{u}_{t-2}, \textbf{u}_{t-1}, \textbf{u}_t)$, are fed into an LSTM network to mitigate directional oscillations. 
All inputs are subsequently processed by multiple fully connected layers with ReLU activation. The critic network $Q(\mathcal{S}_t, \textbf{u}_t)$ outputs a value estimate, while the actor network $\pi (\mathcal{S}_t, \textbf{u}_t)$ generates the next action $\textbf{u}_{t+1}$ for the robot. 
The action consists of the rear wheel’s speed and the front wheel’s angle, $\textbf{u}_{t+1}=(v_{t+1}, \phi_{t+1})$, with the former continuously ranging among the kinetic limitations $v_{t+1} \in [v_{\min}, v_{\max}]$, and the latter continuously ranging between the maximal angles $\phi_t \in [-\phi_{\max}, \phi_{\max}]$. 

The reward function $r_t$ balances efficiency and collision avoidance. To quantify progress toward the goal, the robot’s velocity is projected onto the goal direction as follows:
\begin{equation}
    v_{\parallel} = \textbf{v} \cdot \textbf{p}\textbf{p}^g / |\textbf{p}\textbf{p}^g| \label{parallel}
\end{equation}
The reward function is then defined as:
\begin{align}
    r_t = w_1 v_{\parallel} - w_2 |\phi_t| + w_3 
    \left
    \{\begin{aligned}
        & 0 & d_t>d^s \\
        & d_t - d^s & d^d < d_t < d^s \\
        & -10 & d_t < d^d \\
        & -5 & t > T^{t.o.} \\
        & 5 & \textbf{p}_t = \textbf{p}^g
    \end{aligned}
    \right.
    \label{reward}
\end{align}
where all weight coefficients $w_i$ are positive. The first two terms encourage the robot to reach the goal efficiently while maintaining a stable orientation. In the last term, $d_t$ denotes the distance between the robot and the nearest obstacle, while $d^s$ and $d^d$ denote the safe and dangerous distance thresholds, respectively. $T^{t.o.}$ defines the maximum episode duration. The episode terminates if the robot collides, exceeds the time limit, or successfully reaches the goal.

As reinforcement learning-based exploration alone often leads to frequent collisions, resulting in overly conservative policies where the robot remains stationary, keyboard-controlled demonstrations are used to generate positive training instances. The collected imitation learning data, along with subsequent exploration experiences, is stored in a memory buffer to improve the policy network $\pi (\mathcal{S}_t, \textbf{u}_t)$ and the value function $Q(\mathcal{S}_t, \textbf{u}_t)$.

\section{Simulation and Experiment} \label{Sim_Exp}
\subsection{Training Setup}
\begin{figure}[h]
  \includegraphics[width=1.0\linewidth]{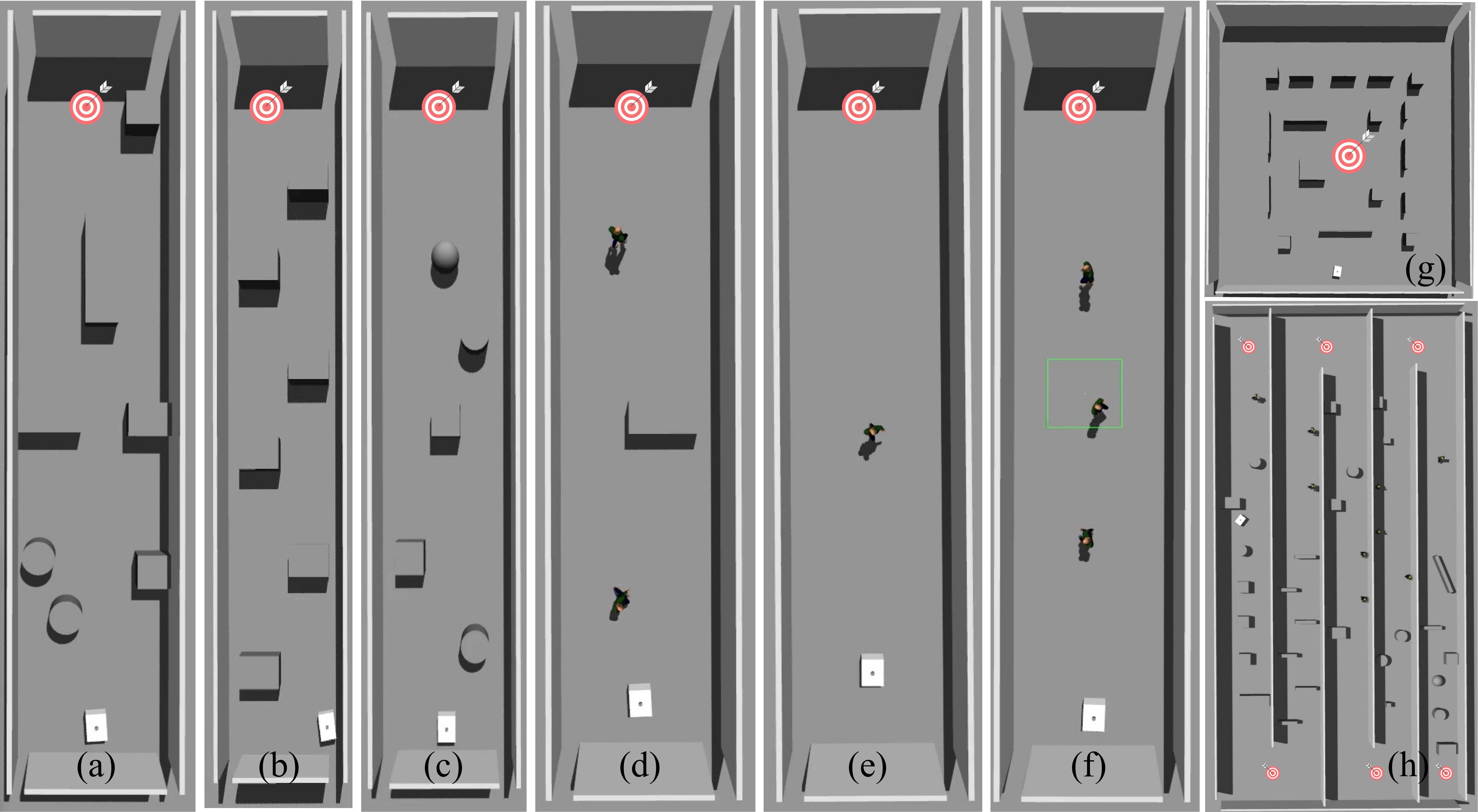}
	\caption{Training scenarios. For more details about the training setup, parameters, and performance of NavG, visit: \href{https://navg-dev.github.io}{https://navg-dev.github.io}. 
  }
\label{fig_simulation}
\end{figure}

The simulation is conducted in Gazebo on Ubuntu 20.04, incorporating rectangular, circular, and elongated obstacles, as well as dynamic pedestrians.
As illustrated in Fig.\ref{fig_simulation}, the simulation environment consists of eight pre-designed scenarios. While the positions and shapes of static obstacles are randomly varied, the velocities and positions of pedestrians continuously change during training. 
To prevent infeasible trajectories caused by entirely random obstacle placements, the following constraints are imposed on the randomness. In corridor scenarios with only static obstacles in Fig.\ref{fig_simulation}(a–c) and those involving dynamic pedestrians in Fig.\ref{fig_simulation}(d–f), the corridor width is randomly adjusted between 4 m and 6 m. The obstacle sizes are randomly scaled between 0.8$\times$ and 1.2$\times$, and their lateral positions (perpendicular to the direction along which the corridor extends) are randomized while ensuring they do not collide with walls or other obstacles. Pedestrian speeds range from 0.3 m/s to 1.5 m/s, and they are continuously moving to provide varying initial positions when the robot enters the scenario. The robot’s starting positions and goals are randomly placed on opposite sides of the corridor. 
In the lobby scenario in Fig.\ref{fig_simulation}(g), obstacle sizes are randomly varied between 0.8$\times$ and 1.2$\times$, and their positions are shifted randomly by up to $\pm$1 m along the long axis. The robot is initialized at one of the four sides, with the target positioned at the center of the scenario. 
In the maze scenario in Fig.\ref{fig_simulation}(h), obstacle sizes remain fixed, but their positions are randomly shifted laterally. Pedestrian speeds range from 0.3 m/s to 1.5 m/s. 
The eight scenarios are visited iteratively in sequence, with obstacle distributions being randomized every three iterations.

The robot has a body shape of [0.824 m × 0.624 m] and is equipped with Ackermann steering, a 16-line Velodyne LiDAR, and an Astra RGB-D camera mounted on top. The front wheel’s speed is constrained within the range of [-0.1 m/s, 1 m/s], while the rear wheel’s steering angle is limited to [$-\pi/4, \pi/4$].
For human detection, four algorithms are alternated after each complete traversal of eight scenarios, including ground-truth data, ground-truth data with Gaussian noise, laser-based HDL-people-tracking~\cite{ref_hdl_tracking}, and camera-based YOLO11m~\cite{Jocher_Ultralytics_YOLO_2023}. 
The use of mixed detection methods increases training complexity while enhancing the robustness in handling perceptual errors with different magnitudes. The Kalman filter is used for human tracking.

The proposed framework, NavG, is implemented in PyTorch and runs on an NVIDIA RTX 3060 GPU. Initially, a single robot is manually controlled across 24 scenarios, after which four robots explore the environment in parallel to generate training data. Over 3000 iterations, the exploration probability decreases linearly from 0.5 to 0.1. The network converges after 8.4 hours, corresponding to 3900 iterations across 31200 scenarios.
Key hyperparameters include a learning rate of $L_r = 3\cdot 10^{-4}$, a discount factor of $\gamma = 0.95$, a batch size of $b_s = 128$, and the Adam optimizer~\cite{ref_adam}. More parameter details can be found in the source code.

Four evaluation metrics are considered: the success rate of reaching the goal $Success$, the average time to reach the goal in successful cases $Time_{success}$, and success weighted by time length $STL$~\cite{francis2023principles}, which accounts for the actual time when the goal is reached successfully and assigns a preset maximum time in the event of a collision, and the ratio of the robot detouring pedestrians from behind $Behind$.
The proposed NavG is compared against four methods: the reinforcement learning-based navigation DRL-VO~\cite{drlvo}, the optimization-based navigation STC-TEB~\cite{stcteb}, NavG without guidance points NavG/P, and NavG without both guidance points and the designed perception-to-planning mapping strategy NavG/PM.
Each scenario in Fig.\ref{fig_simulation} is repeated 50 times with randomly placed obstacles and pedestrians, with half of the tests using HDL for human detection and the other half using YOLO11. The results from scenarios (a, b, c, g) are grouped as static scenarios, while those from (d, e, f, h) are categorized as dynamic scenarios in Table.\ref{table}.

\subsection{Qualitative and Quantitative Analysis in Simulation}

\begin{figure}[t]
  \includegraphics[width=1.0\linewidth]{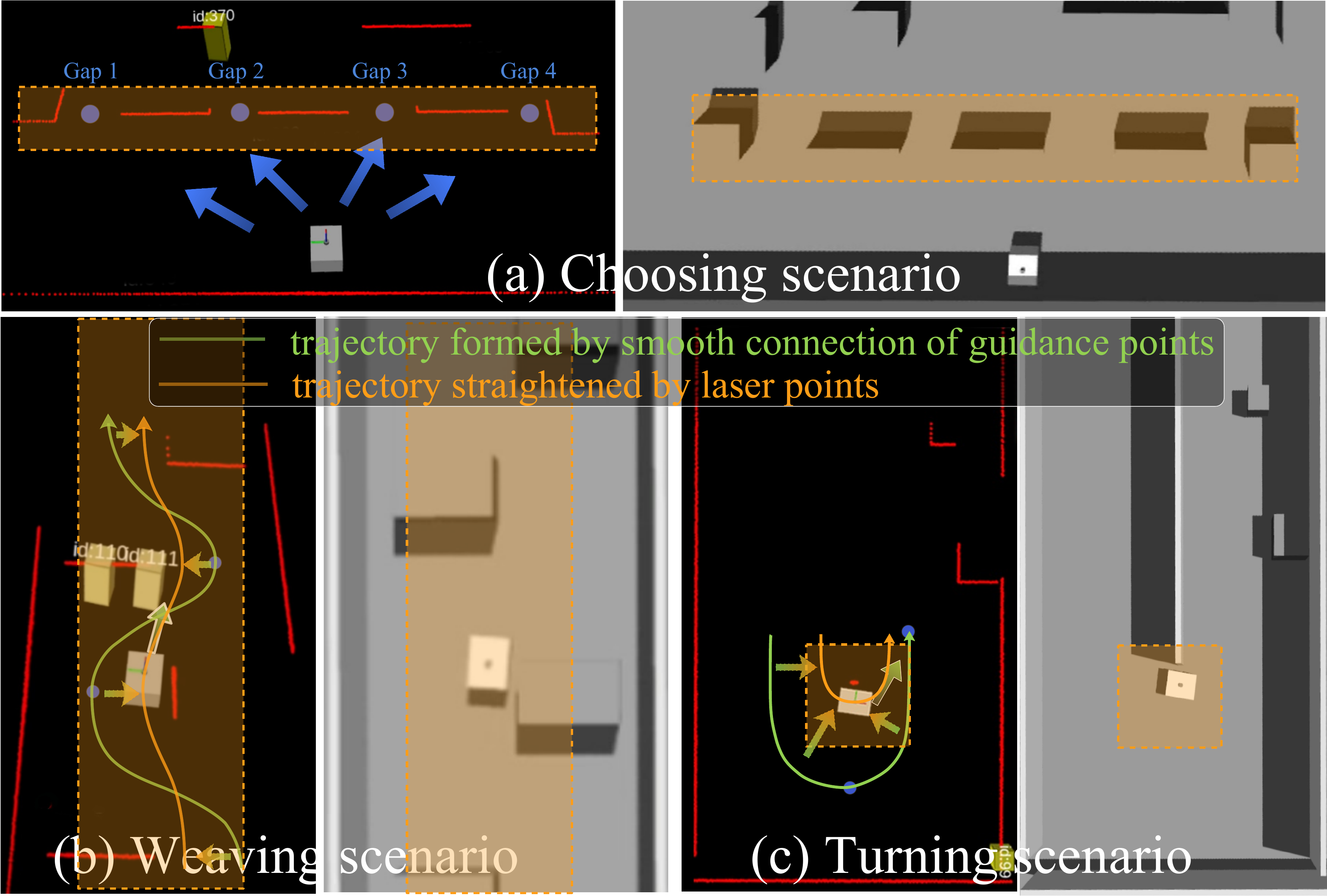}
	\caption{Illustration of the efficient navigation with the help of guidance points and sparse laser. }
	\label{fig_guidance_laser}
\end{figure}

\textbf{Guidance Points \& Sparse Laser for Efficiency}. 
Fig.\ref{fig_guidance_laser} illustrates the role of guidance points in enhancing navigation efficiency by assisting the robot in comprehending the environment and identifying potential directions.

In the presence of five obstacles, as shown in Fig.\ref{fig_guidance_laser}(a), the robot has four possible gaps to navigate through. However, some gaps are too narrow, while others require significant detours. The introduction of guidance points enables the robot to recognize all four gaps. By incorporating laser data, Gap 2 is discarded as it is too narrow for the robot to pass through. NavG selects Gap 3, allowing the robot to efficiently reach the goal by navigating through narrow passages rather than cautiously moving along the scenario’s edges in search of a larger gap, as observed in the compared methods DRL-VO, NavG/P, and NavG/PM. 
This explains why NavG achieves the shortest time to goal of 24.9 seconds in static scenarios, as shown in Table \ref{table}.

In scenarios requiring the robot to weave through obstacles, as shown in Fig.\ref{fig_guidance_laser}(b), guidance points are positioned at the midpoints between obstacles and walls. However, the initial path formed by smoothly connecting these points may contain unnecessary detours. To enhance efficiency, this path should be implicitly refined through a shrinking process, with laser data providing a safe lower boundary. This refinement allows the robot to move closer to obstacles while ensuring safety, as illustrated by the orange trajectory. Here, the guidance points also act as an attractive force, encouraging the robot to maintain a safe distance from obstacles. As a result, NavG achieves the highest success rate of 96\% in reaching the goal, outperforming other approaches.
This example underscores the significance of the guidance points as well as the designed mapping strategy, enabling the robot to adaptively balance efficiency and safety according to the guidance points and laser data. 

In turning scenarios, as shown in Fig.\ref{fig_guidance_laser}(c), the velocity term in the reward function encourages the robot to turn as quickly as possible to maximize its reward. Guidance points indicate the optimal turning location, while laser data ensures safe boundaries during the maneuver. This outcome is primarily attributed to the velocity component $v_{\parallel}$ in the reward function Eq.\ref{reward}, which effectively directs the robot toward the goal while preventing excessive turning.

\begin{table*}[b]
\centering
\caption{Comparison Results}
\label{table}
\begin{tabular}{lcccccccc}
\hline
            & \multicolumn{4}{c}{Static scenarios (a,b,c,g)}                               & \multicolumn{4}{c}{Dynamic scenarios (d,e,f,h)}                              \\ \cline{2-9} 
            & Time$_{success}$ {[}s{]} & Success {[}\%{]} & STL {[}s{]}   & Behind {[}\%{]} & Time$_{success}$ {[}s{]} & Success {[}\%{]} & STL {[}s{]}   & Behind {[}\%{]} \\ \hline
DRL-VO~\cite{drlvo}      & 26.9                    & 92               & 27.9          & /               & 28.2                    & 78               & 30.8          & 42              \\ \hline
STC-TEB~\cite{stcteb}     & 25.2                    & 96      & 25.0          & /               & \textbf{26.3}           & 72               & 30.1          & 56              \\ \hline
NavG/P (ablation)     & 25.8                    & 90               & 27.2          & /               & 28.0                    & 80               & 30.4          & 54              \\ \hline
NavG/PM (ablation)    & 30.0                    & 84               & 31.6          & /               & 29.5                    & 64               & 33.3          & 44              \\ \hline
NavG (ours) & \textbf{24.9}           & \textbf{96}      & \textbf{25.9} & /               & \textbf{26.6}           & \textbf{96}      & \textbf{27.5} & \textbf{74}     \\ \hline
\end{tabular}
\end{table*}

\begin{figure}[t]
  \includegraphics[width=1.0\linewidth]{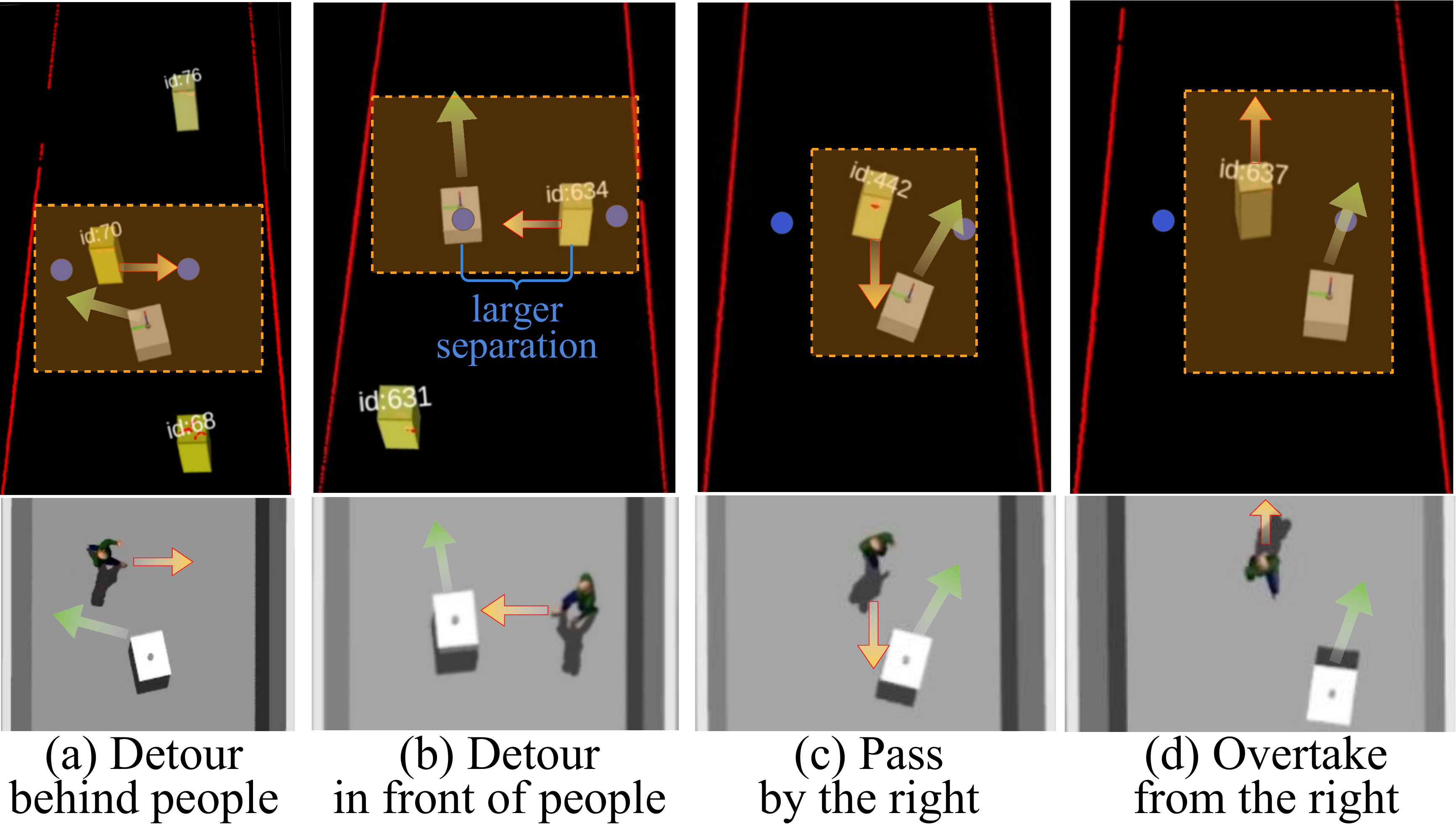}
	\caption{Illustration of the safe navigation with the help of guidance points and human detection.}
	\label{fig_human_guidance}
\end{figure}

\textbf{Guidance Points \& Human Detection for Safety}. 
Fig.\ref{fig_human_guidance} illustrates the role of guidance points in enhancing safety. To rigorously test the robustness of our approach, we employ a suboptimal human detection algorithm, HDL~\cite{ref_hdl_tracking}.
Fig.\ref{fig_human_guidance}(a-b) depicts a scenario in which the robot navigates past a pedestrian. Passing pedestrians in a corridor requires the robot to balance maintaining a safe distance and efficiency. Typically, robots prefer to detour closely around pedestrians to minimize travel time. However, it can be risky if pedestrian speed is inaccurately detected.

Guidance points mitigate this issue in two ways:
First, guidance points explicitly indicate to the RL agent that there are two possible directions to navigate around a pedestrian—either from the front or behind. When the time benefits of both directions are comparable, the agent prefers to detour from behind, as shown in Fig.\ref{fig_human_guidance}(a). This strategy ensures that even if a detection error occurs, the robot can maintain safety by slowing down. The impact of this preference is reflected in the success rate and the ratio of detouring from behind metrics. Among the ablation comparisons of NavG, NavG/P, and NavG/PM, these two metrics show a positive correlation. NavG, which incorporates guidance points, achieves the highest success rate of 92\% with the highest detouring-from-behind ratio of 74\% in Table.\ref{table}.

Second, in cases where detouring in front of a pedestrian offers a clear time advantage, the robot naturally selects that path. However, through collision lessons during training, the agent learns to maintain a larger safety margin, prioritizing safety over minor efficiency gains, as shown in Fig.\ref{fig_human_guidance}(b). As a result, NavG achieves the highest success rate of 96\%, outperforming DRL-VO (78\%) and STC-TEB (72\%) in the dynamic scenarios.
A notable result is that while achieving the highest success rate, NavG maintains an average time to the goal of 26.6 seconds in successful cases, only slightly slower than the fastest method, STC-TEB (26.3 seconds). This efficiency is attributed to the designed mapping strategy, which adaptively selects the optimal safety separation. Furthermore, when considering both successful and failed cases through the STL metric, NavG significantly outperforms all other methods, achieving the highest value of 27.5 seconds.

In other scenarios (c-d), when a pedestrian walks toward or in front of the robot, the guidance points provide the RL agent with two possible navigation options: passing on the left or right. During the imitation learning phase, the human operator instinctively follows social norms, such as passing on the right side (as is common in China). The guidance points facilitate the RL agent’s ability to learn this behavior efficiently, resulting in a preference for (c) passing an oncoming pedestrian from the right and (d) overtaking a pedestrian from the right.

\begin{figure*}
  \includegraphics[width=1.0\linewidth]{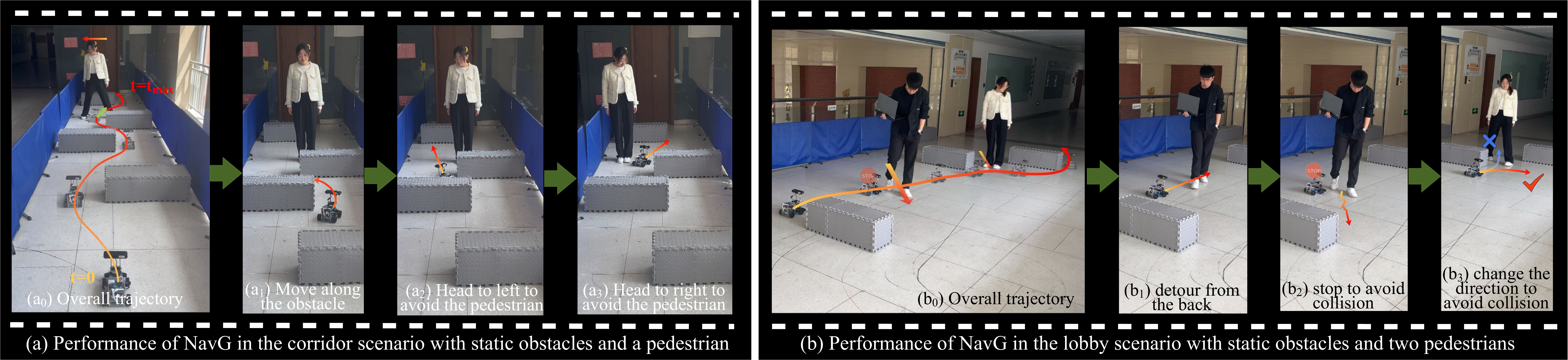}
	\caption{Illustration of the performance of NavG in the corridor and lobby scenarios with obstacles and pedestrians. }
	\label{fig_experiment}
\end{figure*}

\begin{figure}[t]
  \includegraphics[width=1.0\linewidth]{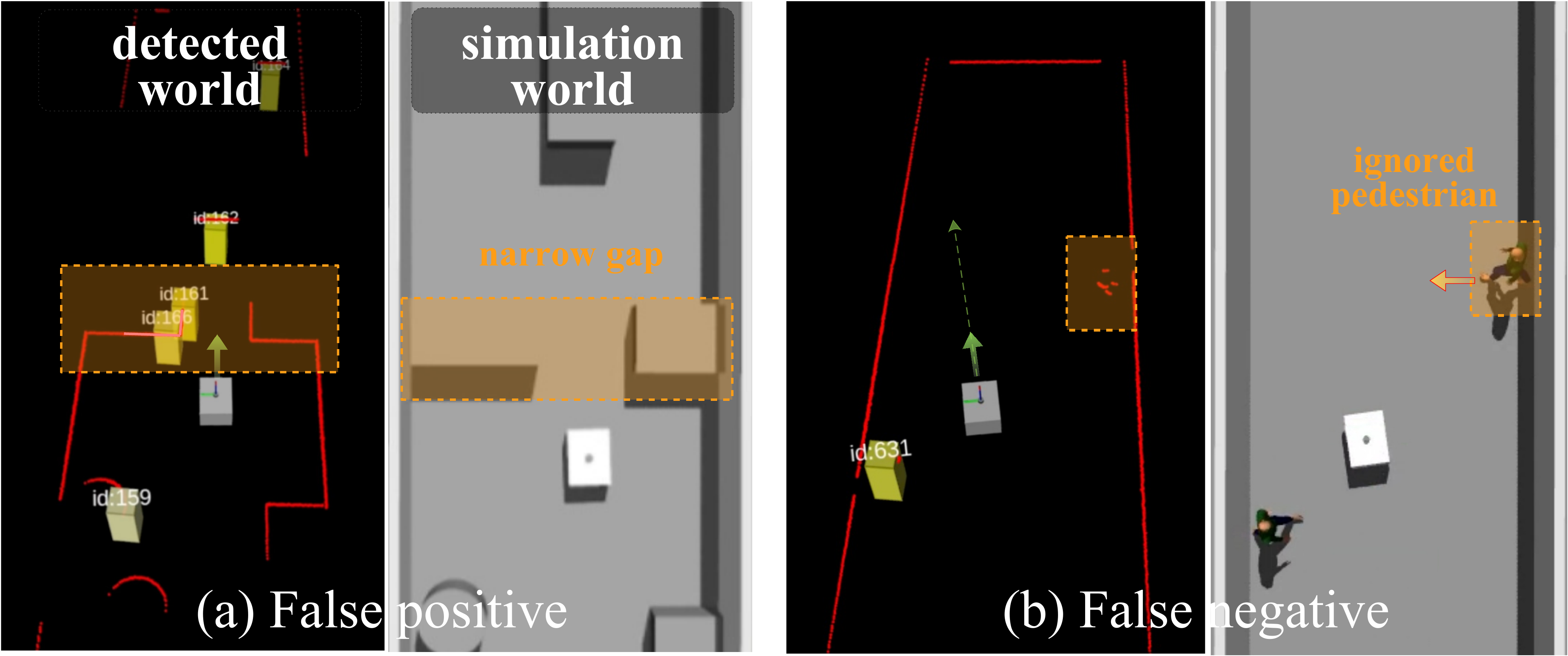}
	\caption{Illustration of the robust navigation with the help of laser information and human detection. }
	\label{fig_laser_human}
\end{figure}

\textbf{Sparse Laser \& Human Detection for Robustness}. 
Fig.\ref{fig_laser_human} illustrates the role of the mapping strategy from perception to planning under a relatively poor human detection algorithm, HDL. Two common issues in human detection include mistaking non-pedestrians for pedestrians (false positives) and failing to detect actual pedestrians (false negatives).
Laser data enables the RL agent to correct these errors. In the false positive example Fig\ref{fig_laser_human}(a), three pedestrians are incorrectly identified. Among them, obstacles 161 and 166 are positioned near a narrow gap, and their circular body shapes further restrict the passage, making navigation more difficult or forcing the robot to maintain unnecessary separation to avoid collisions. However, the linear shape of the surrounding laser points reduces their likelihood of being pedestrians. As a result, our method allows the robot to confidently pass through the gap without being misled by false detections. This capability is reflected in the average time to the goal in successful cases: the time-optimal navigation method STC-TEB requires 25.2 seconds due to extra detours in static scenarios, the ablation method NavG/PM requires 30.0 seconds, whereas our NavG—leveraging the mapping strategy—achieves the lowest time of 24.9 seconds.

In the false negative example Fig\ref{fig_laser_human}(b), when a pedestrian is too close to the wall to be correctly identified, the circular human-like shape in the laser data keeps the robot aware of its presence. In cases of uncertainty, instead of moving straight through the middle of the corridor toward the goal, the robot slightly detours to the left, maintaining a safe margin. This ability allows NavG to achieve the highest success rate of 96\% among all methods.

A notable result in dynamic scenarios is that while NavG does not surpass STC-TEB in terms of average time in successful cases (26.6 seconds vs. 26.3 seconds), this outcome is expected. As a time-optimal method, STC-TEB naturally holds an efficiency advantage in environments with enough traversable space, where perceptual errors have minimal impact on navigation. However, NavG outperforms all other methods across all other metrics, particularly the STL metric, which accounts for both successful and failed cases. We believe this result provides a more comprehensive and convincing demonstration of our approach’s advantages.

\subsection{Hardware Experiment} \label{Exp_Dis}
The experiments are conducted in corridor and lobby scenarios, as shown in Fig.\ref{fig_experiment}.
In the corridor scenario, (a$_1$) the robot initially moves closely along obstacles for efficiency while maintaining an adaptive separation for safety. (a$_2$) Upon detecting a pedestrian standing in its path, the robot naturally steers to the left to avoid her. (a$_3$) When the pedestrian suddenly steps to her right, the robot quickly adjusts by steering right to prevent a collision.
This demonstration highlights NavG’s ability to balance efficiency and safety in the dynamic environment with pedestrians.

In the lobby scenario, (b$_1$) after navigating around an obstacle, the robot initially plans to detour behind a pedestrian dressed in black, considering their relative position and velocity. (b$_2$) However, due to perceptual errors and the pedestrian intentionally stopping, the robot detects insufficient space to pass behind and immediately stops, attempting to reverse to avoid a collision. (b$_3$) Before detecting a pedestrian dressed in white, the robot plans to turn left to navigate through the obstacles. Upon detecting the pedestrian, it quickly adapts by turning right to avoid her.
This demonstration highlights NavG’s robustness in handling unpredictable pedestrian behaviors and perceptual uncertainties.

\section{Conclusion}
This paper focuses on risk-aware navigation in the presence of perceptual errors by introducing guidance points into a reinforcement-learning-based navigation framework. Extracted from the environment, guidance points explicitly suggest potential directions for the robot. To integrate guidance points with other perceptual inputs and correct perceptual errors, we propose a perception-to-planning mapping strategy that enables the RL agent to capture complementary relationships among different sources of perceptual information.
Simulations and real-world experiments conducted in corridors, lobbies, and mazes demonstrate that the proposed NavG achieves the highest success rate and near-optimal travel time. The results show that the robot can confidently navigate through obstacles and robustly interact with pedestrians, even in the presence of perceptual errors.

\bibliographystyle{IEEEtran}
\bibliography{./ref}

\begin{thebibliography}{10}
\providecommand{\url}[1]{#1}
\csname url@samestyle\endcsname
\providecommand{\newblock}{\relax}
\providecommand{\bibinfo}[2]{#2}
\providecommand{\BIBentrySTDinterwordspacing}{\spaceskip=0pt\relax}
\providecommand{\BIBentryALTinterwordstretchfactor}{4}
\providecommand{\BIBentryALTinterwordspacing}{\spaceskip=\fontdimen2\font plus
\BIBentryALTinterwordstretchfactor\fontdimen3\font minus \fontdimen4\font\relax}
\providecommand{\BIBforeignlanguage}[2]{{%
\expandafter\ifx\csname l@#1\endcsname\relax
\typeout{** WARNING: IEEEtran.bst: No hyphenation pattern has been}%
\typeout{** loaded for the language `#1'. Using the pattern for}%
\typeout{** the default language instead.}%
\else
\language=\csname l@#1\endcsname
\fi
#2}}
\providecommand{\BIBdecl}{\relax}
\BIBdecl

\bibitem{ref_review1}
R.~Mirsky, X.~Xiao, J.~Hart, and P.~Stone, ``Conflict avoidance in social navigation—a survey,'' \emph{J. Hum.-Robot Interact.}, vol.~13, no.~1, mar 2024.

\bibitem{ref_review2}
M.~Y. Arafat, M.~M. Alam, and S.~Moh, ``Vision-based navigation techniques for unmanned aerial vehicles: Review and challenges,'' \emph{Drones}, vol.~7, no.~2, 2023.

\bibitem{ref_obj_detection1}
Z.~Zou, K.~Chen, Z.~Shi, and Y.~Guo, ``Object detection in 20 years: A survey,'' \emph{Proceedings of IEEE}, vol. 111, no.~3, pp. 257--276, 2023.

\bibitem{ref_obj_detection2}
T.~Diwan, A.~Ani, and J.~Tembhurne, ``Object detection using yolo: challenges, architectural successors, datasets and applications,'' \emph{Multimedia Tools and Applications}, vol.~82, 08 2022.

\bibitem{sam}
Y.~Zhang, T.~Cheng, R.~Hu, L.~Liu, H.~Liu, L.~Ran, X.~Chen, W.~Liu, and X.~Wang, ``Evf-sam: Early vision-language fusion for text-prompted segment anything model,'' \emph{arXiv preprint arXiv:2406.20076}, 2024.

\bibitem{evora}
X.~Cai, S.~Ancha, L.~Sharma, P.~R. Osteen, B.~Bucher, S.~Phillips, J.~Wang, M.~Everett, N.~Roy, and J.~P. How, ``Evora: Deep evidential traversability learning for risk-aware off-road autonomy,'' \emph{IEEE Transactions on Robotics}, vol.~40, pp. 3756--3777, 2024.

\bibitem{drlvo}
Z.~Xie and P.~Dames, ``Drl-vo: Learning to navigate through crowded dynamic scenes using velocity obstacles,'' \emph{IEEE Transactions on Robotics}, vol.~39, no.~4, pp. 2700--2719, 2023.

\bibitem{khambhaita2020viewing}
H.~Khambhaita and R.~Alami, ``Viewing robot navigation in human environment as a cooperative activity,'' in \emph{Robotics Research: The 18th International Symposium ISRR}.\hskip 1em plus 0.5em minus 0.4em\relax Springer, 2020, pp. 285--300.

\bibitem{stcteb}
Z.~Zhu, Q.~Zhang, Y.~Song, Y.~Yang, and J.~Liu, ``Stc-teb: Spatial-temporally complete trajectory generation based on incremental optimization,'' \emph{IEEE Robotics and Automation Letters}, vol.~10, no.~2, pp. 1289--1296, 2025.

\bibitem{10161222}
Q.~Zhang, Z.~Hu, Y.~Song, J.~Pei, and J.~Liu, ``The human gaze helps robots run bravely and efficiently in crowds,'' in \emph{2023 IEEE International Conference on Robotics and Automation (ICRA)}, 2023, pp. 7540--7546.

\bibitem{9775638}
V.~B. Hoang, V.~H. Nguyen, T.~D. Ngo, and X.-T. Truong, ``Socially aware robot navigation framework: Where and how to approach people in dynamic social environments,'' \emph{IEEE Transactions on Automation Science and Engineering}, vol.~20, no.~2, pp. 1322--1336, 2023.

\bibitem{9982200}
X.~Cai, M.~Everett, J.~Fink, and J.~P. How, ``Risk-aware off-road navigation via a learned speed distribution map,'' in \emph{2022 IEEE/RSJ International Conference on Intelligent Robots and Systems (IROS)}, 2022, pp. 2931--2937.

\bibitem{AgileButSafe}
T.~He, C.~Zhang, W.~Xiao, G.~He, C.~Liu, and G.~Shi, ``Agile but safe: Learning collision-free high-speed legged locomotion,'' in \emph{Robotics: Science and Systems (RSS)}, 2024.

\bibitem{10610665}
A.~Sridhar, D.~Shah, C.~Glossop, and S.~Levine, ``Nomad: Goal masked diffusion policies for navigation and exploration,'' in \emph{2024 IEEE International Conference on Robotics and Automation (ICRA)}, 2024, pp. 63--70.

\bibitem{hirose2024selfi}
N.~Hirose, D.~Shah, K.~Stachowicz, A.~Sridhar, and S.~Levine, ``{SELFI}: Autonomous self-improvement with {RL} for vision-based navigation around people,'' in \emph{8th Annual Conference on Robot Learning}, 2024.

\bibitem{ref_2_3}
C.~Chen, Y.~Liu, S.~Kreiss, and A.~Alahi, ``Crowd-robot interaction: Crowd-aware robot navigation with attention-based deep reinforcement learning,'' in \emph{2019 international conference on robotics and automation (ICRA)}.\hskip 1em plus 0.5em minus 0.4em\relax IEEE, 2019, pp. 6015--6022.

\bibitem{ref_1}
Y.~F. Chen, M.~Liu, M.~Everett, and J.~P. How, ``Decentralized non-communicating multiagent collision avoidance with deep reinforcement learning,'' in \emph{2017 IEEE International Conference on Robotics and Automation (ICRA)}, 2017, pp. 285--292.

\bibitem{ref_2}
T.~Xuan~Tung and T.~Dung~Ngo, ``Socially aware robot navigation using deep reinforcement learning,'' in \emph{2018 IEEE Canadian Conference on Electrical and Computer Engineering (CCECE)}, 2018, pp. 1--5.

\bibitem{ref_3}
Z.~Hu, Y.~Zhao, S.~Zhang, L.~Zhou, and J.~Liu, ``Crowd-comfort robot navigation among dynamic environment based on social-stressed deep reinforcement learning,'' \emph{International Journal of Social Robotics}, vol.~14, no.~4, pp. 913--929, 2022.

\bibitem{ref_4}
M.~Everett, Y.~F. Chen, and J.~P. How, ``Motion planning among dynamic, decision-making agents with deep reinforcement learning,'' in \emph{2018 IEEE/RSJ International Conference on Intelligent Robots and Systems (IROS)}.\hskip 1em plus 0.5em minus 0.4em\relax IEEE, 2018, pp. 3052--3059.

\bibitem{ref_5}
M.~\vspace{0mm}Everett, Y.~F. Chen, and J.~P. How, ``Collision avoidance in pedestrian-rich environments with deep reinforcement learning,'' \emph{IEEE Access}, vol.~9, pp. 10\,357--10\,377, 2021.

\bibitem{ref_9}
A.~Thomas, G.~Ferro, F.~Mastrogiovanni, and M.~Robba, ``Computational tradeoff in minimum obstacle displacement planning for robot navigation,'' in \emph{2023 IEEE International Conference on Robotics and Automation (ICRA)}, 2023, pp. 3635--3641.

\bibitem{roth2024viplannervisualsemanticimperative}
P.~Roth, J.~Nubert, F.~Yang, M.~Mittal, and M.~Hutter, ``Viplanner: Visual semantic imperative learning for local navigation,'' 2024.

\bibitem{ref_21}
M.~Lodel, B.~Brito, A.~Serra-G\'{o}mez, L.~Ferranti, R.~Babu\v{s}ka, and J.~Alonso-Mora, ``Where to look next: Learning viewpoint recommendations for informative trajectory planning,'' in \emph{2022 International Conference on Robotics and Automation (ICRA)}, 2022, pp. 4466--4472.

\bibitem{ref_19}
F.~Tsang, T.~Walker, R.~A. MacDonald, A.~Sadeghi, and S.~L. Smith, ``Lamp: Learning a motion policy to repeatedly navigate in an uncertain environment,'' \emph{IEEE Transactions on Robotics}, vol.~38, no.~3, pp. 1638--1652, 2022.

\bibitem{ref_20}
X.~Xiao, Z.~Wang, Z.~Xu, B.~Liu, G.~Warnell, G.~Dhamankar, A.~Nair, and P.~Stone, ``Appl: Adaptive planner parameter learning,'' \emph{Robotics and Autonomous Systems}, vol. 154, p. 104132, 2022.

\bibitem{ref_8}
R.~Cimurs, I.~H. Suh, and J.~H. Lee, ``Goal-driven autonomous exploration through deep reinforcement learning,'' \emph{IEEE Robotics and Automation Letters}, vol.~7, no.~2, pp. 730--737, 2021.

\bibitem{ref_16}
K.~Weerakoon, A.~J. Sathyamoorthy, U.~Patel, and D.~Manocha, ``Terp: Reliable planning in uneven outdoor environments using deep reinforcement learning,'' in \emph{2022 International Conference on Robotics and Automation (ICRA)}, 2022, pp. 9447--9453.

\bibitem{ref_22}
E.~Kaufmann, L.~Bauersfeld, A.~Loquercio, M.~M{\"u}ller, V.~Koltun, and D.~Scaramuzza, ``Champion-level drone racing using deep reinforcement learning,'' \emph{Nature}, vol. 620, no. 7976, pp. 982--987, 2023.

\bibitem{ref_voronoi}
B.~Lau, C.~Sprunk, and W.~Burgard, ``Improved updating of euclidean distance maps and voronoi diagrams,'' in \emph{2010 IEEE/RSJ International Conference on Intelligent Robots and Systems}, 2010, pp. 281--286.

\bibitem{ref_lstm}
S.~Hochreiter and J.~Schmidhuber, ``Long short-term memory,'' \emph{Neural Computation}, vol.~9, no.~8, pp. 1735--1780, 1997.

\bibitem{ref_10}
T.~Haarnoja, A.~Zhou, P.~Abbeel, and S.~Levine, ``Soft actor-critic: Off-policy maximum entropy deep reinforcement learning with a stochastic actor,'' in \emph{Proceedings of the 35th International Conference on Machine Learning}, ser. Proceedings of Machine Learning Research, J.~Dy and A.~Krause, Eds., vol.~80.\hskip 1em plus 0.5em minus 0.4em\relax PMLR, 10--15 Jul 2018, pp. 1861--1870.

\bibitem{ref_hdl_tracking}
K.~Koide, J.~Miura, and E.~Menegatti, ``A portable three-dimensional lidar-based system for long-term and wide-area people behavior measurement,'' \emph{International Journal of Advanced Robotic Systems}, vol.~16, no.~2, p. 1729881419841532, 2019.

\bibitem{Jocher_Ultralytics_YOLO_2023}
\BIBentryALTinterwordspacing
G.~Jocher, J.~Qiu, and A.~Chaurasia, ``{Ultralytics YOLO},'' Jan. 2023. [Online]. Available: \url{https://github.com/ultralytics/ultralytics}
\BIBentrySTDinterwordspacing

\bibitem{ref_adam}
T.~Campbell, M.~Liu, B.~Kulis, J.~P. How, and L.~Carin, ``Dynamic clustering via asymptotics of the dependent dirichlet process mixture,'' in \emph{Advances in Neural Information Processing Systems}, vol.~26.\hskip 1em plus 0.5em minus 0.4em\relax Curran Associates, Inc., 2013.

\bibitem{francis2023principles}
A.~Francis, C.~P{\'e}rez-d'Arpino, C.~Li, F.~Xia, A.~Alahi, R.~Alami, A.~Bera, A.~Biswas, J.~Biswas, R.~Chandra \emph{et~al.}, ``Principles and guidelines for evaluating social robot navigation algorithms,'' \emph{arXiv preprint arXiv:2306.16740}, 2023.

\end{thebibliography}

\end{document}